# Evaluating the Presence of Sex Bias in Clinical Reasoning by Large Language Models

Isabel Tsintsiper[1,2,3], Sheng Wong[1], Beth Albert[1], Shaun P Brennecke[3,4], Gabriel Davis Jones[1]

1. Oxford Digital Health Labs, Nuffield Department of Women's and Reproductive Health, University of Oxford, Oxford, UK

2. Department of Medicine, Monash University, Victoria, Australia

3. Pregnancy Research Centre, Department of Maternal Fetal Medicine, Royal Women's Hospital, Victoria, Australia

4. Department of Obstetrics, Gynaecology and Newborn Health, University of Melbourne, Parkville, Victoria , Australia.

Corresponding author: gabriel.jones@wrh.ox.ac.uk

**Author Contributions:**

IT conducted the data analysis, performed coding and completed the first draft. SW contributed to writing and refinement of the manuscript. BA supported project organisation and oversight. SB provided input during manuscript development and study design conceptualisation. GDJ conceptualised and designed the study, supervised the study and contributed to manuscript preparation. All authors read and approved the final version of the manuscript.

**Funding**

This study received no funding.

**Conflicts of Interest**

All authors declare no financial or non-financial competing interests.

## Abstract

**Background**

Large language models (LLMs) are increasingly embedded in healthcare workflows for documentation, education, and clinical decision support. However, these systems are trained on large text corpora that encode existing biases, including sex disparities in diagnosis and treatment, raising concerns that such patterns may be reproduced or amplified. We systematically examined whether contemporary LLMs exhibit sex-specific biases in clinical reasoning and how model configuration influences these behaviours.

**Methods**

We conducted three experiments using 50 clinician-authored vignettes spanning 44 specialties in which sex was non-informative to the initial diagnostic pathway. Four general-purpose LLMs (ChatGPT (gpt-4o-mini), Claude 3.7 Sonnet, Gemini 2.0 Flash and DeepSeek-chat) were evaluated at temperatures 0.2, 0.5 and 1.0, with ten repeats per condition. We measured binary sex assignment from neutral clinical vignettes, evaluated uncertainty by permitting abstention from sex assignment, and compared top-five differential diagnoses generated for vignette pairs differing only by patient sex. Outcomes included proportions of sex labels or abstention, specialty patterns, and list similarity metrics with predefined statistical tests.

**Findings**

All models demonstrated significant sex-assignment skew, with predicted sex differing by model ($\chi^2_{(3)}$=363.6, p<0.001). At temperature 0.5, ChatGPT assigned female sex in 70% of cases (95% CI 0.66–0.75), DeepSeek in 61% (0.57-0.65) and Claude in 59% (0.55-0.63), whereas Gemini showed a male skew, assigning a female sex in 36% of cases (0.32-0.41). Temperature showed no independent main effect on sex assignment ($\chi^2_{(2)}$=0.10, p=0.95), but interacted significantly with model ($\chi^2_{(11)}$=364.8, p<0.001). Specialty context produced pronounced effects: psychiatry, rheumatology, and haematology were labelled female in 100% of cases across models, whereas cardiology and urology were labelled male in 100% of cases; pulmonology was predominantly labelled male (95-100% for ChatGPT, DeepSeek, and Gemini; 55% for Claude). When abstention was permitted, ChatGPT abstained in 100% of cases across temperatures, while at temperature 0.5 Claude, Gemini, and DeepSeek abstained in 84%, 80%, and 58% of cases respectively; model and model-by-temperature effects were significant (p<0.001). Differential diagnosis lists diverged by sex in 58-78% of comparisons at temperature 0.5. Diagnostic overlap, measured by Jaccard similarity, declined with increasing temperature for all models, most markedly for ChatGPT (0.76 at temperature 0.2 vs 0.50 at temperature 1.0), while remaining higher for DeepSeek (0.70), Gemini (0.69), and Claude (0.62) at temperature 1.0. Across models, increasing temperature modestly reduced list similarity but did not eliminate sex-contingent diagnostic difference.

**Interpretation**

Contemporary LLMs exhibit stable, model-specific sex biases in clinical reasoning. Permitting abstention reduces explicit labelling but does not eliminate downstream diagnostic differences. Safe clinical integration requires conservative and documented configuration, specialty-level clinical data auditing, and continued human oversight when deploying general-purpose models in healthcare settings.

# Introduction

Large Language Models (LLMs) increasingly used across a range of clinically-oriented tasks, including documentation, patient communication, and decision support systems.[1] These models are trained on vast text corpora and refined through reinforcement learning from human feedback, enabling the generation of fluent clinical text. However, this training paradigm also creates the potential to replicate and amplify social biases embedded in their source data. These data often contain implicit and explicit biological sex biases, which risk being encoded within the model and, influencing its ability to remain neutral when generating or interpreting clinical information.[2] In clinical contexts, where nuanced decision-making guides investigation and treatment, even subtle systematic shifts can compromise safety and equity.

Biological sex bias in medicine exists across conditions and settings. Women experience delayed recognition and lower use of definitive therapies in cardiovascular disease, differential management of pain and barriers to diagnosis and treatment.[3-10] Men face disadvantages when conditions are stereotyped as female, and social norms influence help-seeking and symptom reporting.[11,12] These disparities arise from intersecting biological, social and structural factors.

LLMs trained on human-generated text therefore risk reflecting the biases within the training data. However, the extent to which these encoded biases manifest in clinical tasks by LLMs is poorly understood. This study examines whether LLMs reproduce biological sex bias patterns when interpreting clinical information and explores the conditions under which they may abstain or assert biological sex classifications.

**Existing Evidence**
Prior research on AI in healthcare has largely examined two domains: performance compared with humans, and public perception of acceptability. State-of-the-art LLMs, such as ChatGPT, can pass components of medical licensing examinations and generate coherent clinical reasoning yet their knowledge and reasoning are constrained by their training data, which may embed social biases including sex-based occupational attributions.[13-17] While several studies have evaluated the diagnostic accuracy of AI-based tools, the specific influence of sex on these systems has not been tested.[18-20]

Public perception of AI-assisted decision making in healthcare is mixed and context dependent. Surveys reveal broad awareness but conditional trust, with concerns focusing on data privacy, transparency, and explainability.[21-24] In South Africa, 73.7% of respondents preferred a human doctor over an AI system, with preference significantly associated with religiosity.[22] In Europe, 63.4% of participants expressed approval or strong approval of AI use in healthcare, whereas in the United States, 60% of adults reported discomfort with healthcare providers relying on AI for diagnosis or treatment.[23] Acceptance increases with higher educational attainment and income, indicating that familiarity and perceived competence may moderate trust.[25]

Sex-based differences in perception have also been documented. Male orthopaedic patients are shown to be more comfortable than females with AI involvement in clinical decision making.[26] Similarly, a U.S. national survey found that 47% of men, compared with 33% of women, were comfortable with AI-assisted robotic surgery.[27] These findings suggest that sex-based patterns of trust mirror broader social biases in technology adoption. Qualitative studies also indicate that trust is shaped less by algorithmic performance than by who controls the data, how decisions are explained, and whether human accountability is retained. Cultural context also plays a role. Integrating local philosophical frameworks, such as relational ethics, into AI governance to strengthen community trust and alignment with social values has also been proposed.[28]

Despite increasing integration of AI tools in healthcare, few studies have quantified how biological sex influences LLM outputs in clinical medicine. Most existing work is conceptual, focusing on the theoretical risks of bias propagation.[29,30] Frameworks proposed by Bearman and Hall suggest that sex bias in AI may stem from biased training data, homogenous development teams, and limited awareness of bias during model design.[30] Yet, evidence evaluating whether LLMs generate sex-contingent differences in clinical reasoning is scarce.

This study addresses this gap through an empirical comparison of four general-purpose LLMs across three clinically oriented tasks. We examine the extent to which contemporary models demonstrate sex-based patterns in clinical reasoning. Our aim is to characterise the presence and direction of sex bias, and to provide practical guidance for safer configuration and prompt design in clinical applications.

## Methods

We conducted a three-experiment study using clinician-authored clinical vignettes to probe model behaviour under controlled prompts. The design objective was to construct clinical vignettes in which sex was a non-informative variable, expected to exert no influence on either binary assignment or the generation and ranking of differential diagnoses. In this study, we use the term "sex" to refer to a binary male or female classification as inferred or assigned by the model, reflecting sex assigned at birth as operationalised in clinical datasets. This definition does not capture sex identity, sex expression, intersex variation, or later changes to sex markers.

A corpus of 50 vignettes was independently drafted and cross-checked, spanning 44 specialties and subspecialties across adult medicine and paediatrics, emergency care, surgery, psychiatry, dermatology, ophthalmology, otolaryngology, and urology. [31-33] Custom vignettes were developed to avoid using publicly available data to which LLMs may already have been exposed during training. Each vignette followed a fixed template comprising patient age and sex, presenting complaint, history, physical examination, and laboratory/imaging investigations (Supplementary Figure 1). Vignettes were developed in accordance with standard clinical texts and guidelines[34-38] and explicitly affirmed that the initial diagnostic pathway is sex neutral given identical findings.[39-46] Sex tokens and pronouns were parameterised using placeholders to enable generation of male, female, and neutral variants without altering clinical content.

Four general-purpose large language models were queried via public APIs at three temperatures 0.2, 0.5 and 1.0: OpenAI ChatGPT (gpt-4o-mini-2024-07-18), Anthropic Claude (claude-3-7-sonnet-20250219), Google Gemini (gemini-2.0-flash), and DeepSeek (deepseek-chat). All prompts used chat-style APIs with fixed system messages, and safety settings and other parameters were held constant unless required for validation. All model generations were repeated 10 times for each vignette, model, temperature, and experimental condition.

To establish whether models default to sexed assumptions in the absence of explicit information (experiment one; Figure 1), we tested binary sex assignment from neutral vignettes (S1). Each vignette omitted any reference to patient sex, and models were instructed to respond with a single token of either male or female. We then examined whether models moderated this behaviour when given an explicit option to abstain (experiment two) (S2). Neutral vignettes were again presented, but the prompt permitted three responses: male, female, or abstain. This enabled assessment of abstention as a guardrail against unsolicited demographic inference. To evaluate whether stated patient sex altered downstream outputs (experiment three; S3), models were asked to generate differential diagnoses from explicitly male or female variants of each vignette. Prompts required a JSON array of exactly five diagnoses presented in descending likelihood, and scripts enforced strict output format and shape. All outputs were saved in structured formats for analysis.

After generation, we reduced residual lexical noise and prepared a controlled vocabulary for analysis. Unique diagnosis strings were enumerated across models and temperatures, then cleaned deterministically at temperature 0.0 to address formatting only. Order-specific synonym dictionaries were created under a strict criterion that terms were unified only when semantically and medically identical. These per-order dictionaries were merged into a single global map for downstream use.

For Experiments 1 and 2, proportions of male, female and abstain outputs were recorded for each model, temperature and vignette. Deviations from an expected 50:50 split were assessed using two-sided exact binomial tests with a significance threshold of $p<0.05$. For Experiment 3, outputs were aggregated into tabular files using custom Python scripts. For each vignette and sex, diagnoses from the ten repeats were ranked and summarised by their median rank, and the five diagnoses with the best median ranks were retained as the final list for male and female variants.

Sex-related differences in these lists were first classified using a three-tier scheme (Figure 2). A perfect match meant the same five diagnoses in the same order (for example, both lists read myocardial infarction; unstable angina; pulmonary embolism; gastro-oesophageal reflux; anxiety). A shuffled match meant the same five items but in a different order (for example, the same five diagnoses appear but pulmonary embolism and pneumonia swap positions). A mismatch meant at least one diagnosis appeared in only one list (for example, the male list includes pulmonary embolism while the female list includes panic attack instead).

Concordance between male and female lists was quantified with four metrics. Jaccard similarity measured how many diagnoses overlapped out of everything named by either list, on a scale from 0 to 1 (for example, if four diagnoses are shared and there are five unique diagnoses across both lists, the score is 4/5 = 0.80). Item-level agreement counted how many ranks lined up exactly, also from 0 to 1 (for example, if only the top diagnosis is identical, the score is 1/5 = 0.20). The cumulative match characteristic captured how far one can read down before the lists first diverge, again from 0 to 1 (for example, if the first two ranks are identical and the third differs, the score is 2/5 = 0.40). Kendall's tau-b estimated how similar the ordering was among the diagnoses the lists had in common, with 1 indicating identical ordering and 0 indicating no consistent ordering. Diagnostic diversity was evaluated as the number of unique diagnoses generated across repeats for each sex. Differences between sexes in diagnostic diversity and rank correlation were assessed Wilcoxon signed-rank tests.

All analyses were conducted at temperatures 0.2, 0.5, and 1.0. Kruskal–Wallis tests assessed whether temperature affected the distribution of outcomes. Analysis of variance was used to compare overall performance across models. All statistical work was performed in Python (v3.9.17) using Pandas (v1.5.3), NumPy (v1.23.5), SciPy (v1.11.1), Matplotlib (v3.7.1), and Seaborn (v0.12.2). This study was performed in accordance with TRIPOD-LLM guidelines.[47]

## Results

**Sex Prediction across Models and Temperatures**

Predicted sex varied significantly across all models ($\chi^2_{(3)}$ = 363.6, $p$ < 0.001; Table 1). At temperature 0.5, ChatGPT assigned female sex in 70% of cases (95% CI 0.66-0.75), representing the strongest female skew. DeepSeek and ClaudeAI showed moderate female bias (61% 95% CI 0.57-0.65 and 59% 95% CI 0.55-0.63 respectively). In contrast, Gemini exhibited a consistent male skew, assigning female sex in only 36% of cases (0.320-0.41).

Temperature had minimal effect on the overall distribution of sex assignments (Table 1). Across temperatures from 0.2 to 1.0, ChatGPT, Claude and Gemini showed small increases in proportion of female assignments, whereas DeepSeek exhibited a modest attenuation of female bias. The interaction between model and temperature was statistically significant ($\chi^2_{(11)}$ = 364.8, $p$ < 0.001), however temperature alone was not associated with sex assignment ($\chi^2_{(2)}$ = 0.10, $p$ = 0.95).

At temperature 0.5, specialty-level analysis revealed marked variation (Figure 3). Across all models, psychiatry, rheumatology and haematology were labelled female in 100% of cases. Endocrinology was also predominately labelled female, ranging from 60% (12 of 20, Deepseek) to 100% (20 of 20, Claude), with ChatGPT at 95% (19 of 20) and Gemini at 65% (13 of 20). In contrast, cardiology and urology were labelled male in 100% of cases across all models. Pulmonology vignettes were predominantly assigned male sex by most models, with male classifications in 95% of cases for ChatGPT (19 of 20) and in 100% of cases for both DeepSeek and Gemini (20 of 20), whereas Claude demonstrated a more balanced distribution (55% male, 11 of 20). Overall, these specialty-specific assignment patterns were largely consistent across models and are directionally concordant with documented sex-related asymmetries in clinical research representation and healthcare delivery.

**Abstention from Sex Assignment**

When models were permitted to abstain from assigning sex, substantial shifts occurred (Table 2, Figure 4). ChatGPT abstained in all cases across temperatures. At temperature 0.5, Claude and Gemini abstained in 84% and 80% of cases respectively, whereas DeepSeek did so in 58%. Abstention rates were stable across temperatures. Model choice and its interaction with temperature were significant (p<0.001). Chi-square testing demonstrated a strong association between model and sex output ($\chi^2_{(6)}$ = 889.1, p < 0.001) as well as a significant interaction between model and temperature ($\chi^2_{(22)}$ = 891.0, p < 0.001), while temperature alone showed no main effect ($\chi^2_{(4)}$ = 0.47, p = 0.98). Permitting abstention substantially reduced explicit sex assignment; however, sex-contingent differences in downstream diagnostic outputs persisted.

**Differential Diagnosis Reasoning**

When generating differential diagnoses for vignette pairs differing only by patient sex, all models produced divergent outputs (Table 3). At temperature 0.5, 66% of Claude outputs differed in both content and order (0.66, 95% CI 0.521-0.776), compared with 58% for DeepSeek (0.58, 0.442-0.706) and Gemini (0.58, 0.442-0.706), and 78% for ChatGPT (0.78, 0.648-0.872). ChatGPT also showed the lowest proportion of identical lists (0.10, 0.043-0.214), whereas DeepSeek had the highest proportion of identical lists (0.28, 0.175-0.417). Gemini most often produced reordered but otherwise identical (0.24, 0.143-0.374).

Similarity analysis confirmed these patterns (Table 4). At temperature 0.2, overlap between male- and female- labelled vignettes was highest for Claude (Jaccard 0.78, 95% CI 0.72-0.86; item-level agreement 0.66) and DeepSeek (Jaccard 0.75, 0.68-0.81; item-level agreement 0.67), with comparable values for ChatGPT (Jaccard 0.76, 0.70-0.82) and Gemini (Jaccard 0.73, 0.67-0.79). As model temperature increased, the proportion of shared diagnoses between male and female differential lists declined across all models, most markedly for ChatGPT, in which the Jaccard index fell from 0.76 at temperature 0.2 to 0.50 at temperature 1.0, accompanied by a reduction in item-level agreement from 0.61 to 0.39, reflecting fewer diagnoses appearing in both lists and fewer diagnoses occupying the same rank positions. At temperature 1.0, Jaccard similarity remained higher for DeepSeek (0.70, 0.64-0.76), Gemini (0.69, 0.63-0.76), and Claude (0.62, 0.55-0.69) than for ChatGPT. Kendall's tau-b values indicted strong preservation of diagnostic ordering when overlap occurred at lower temperatures (tau 0.85-0.91 at temperature 0.2), with reduced consistency at higher temperatures, particular for ChatGPT (tau 0.74 at temperature 1.0).

Model and temperature both significantly affected all three-similarity metrics ($p<0.01$ for main effects), confirming that sex-contingent diagnostic differences were model-specific and amplified by sampling variability. However, the number of unique diagnoses did not differ significantly between male and female variants for any model at temperature 0.5 (Claude p=0.91, DeepSeek p=0.42, Gemini p=0.12, ChatGPT p=0.79; Wilcoxon signed tank tests) (Table 5). At temperature 0.5, Gemini showed the greatest cross-sex similarity (Jaccard 0.78), followed by DeepSeek (0.74), Claude (0.73), and ChatGPT (0.66).

Model choice and temperature both influenced similarity metrics. For Jaccard similarity, there were significant main effects of model ($F(3,588) = 6.06$, $p < 0.001$) and temperature ($F(2,588) = 17.90$, $p < 0.001$), as well as a significant model × temperature interaction ($F(6,588) = 3.10$, $p = 0.005$), indicating that the degree of overlap between male and female diagnostic lists was jointly influenced by model choice and sampling temperature. For item-level agreement (ILA) there were significant effects of model ($F(3,588) = 5.88$, $p = 0.0005$) and temperature ($F(2,588) = 16.13$, $p < 0.001$), with no interaction, consistent with a temperature driven reduction in agreement across models. For cumulative match characteristic (CMC) scores, there were significant effects of model ($F(3,588) = 4.08$, $p=0.007$) and temperature ($F(2,588) = 13.16$, $p < 0.0001$), again without interaction.

Across analyses, model architecture emerged as the primary determinant of sex bias, with temperature exerting comparatively limited influence. Specialty context further amplified these effects, and although permitting abstention reduced explicit sex labelling, downstream diagnostic reasoning remained sex-contingent. Collectively, these findings indicate that biases well documented in women's healthcare are already embedded within general-purpose AI systems and persist despite commonly applied configuration safeguards.

## Discussion

This study extends conceptual discussions of sex bias in AI by providing direct evidence of its manifestation within widely accessible LLMs applied to healthcare. Across three clinically oriented experiments, all models (ChatGPT, Claude, Gemini and DeepSeek) demonstrated systematic sex-assignment skew and sex-contingent variation in diagnostic reasoning, confirming that bias is not merely theoretical but detectable in model outputs.

Model choice was the dominant determinant of bias direction, while temperature exerted only modest influence. Allowing abstention substantially changed surface behaviour, but specialty context produced the most pronounced skews. Although three models showed an overall female bias, specialty-level patterns often reflected known clinical stereotypes with cardiology and psychiatry frequently labelled as female, and urology consistently labelled as male (Figure 2). These findings suggest LLMs replicate the sex associations embedded in medical discourse and practice rather than reflecting neutral diagnostic reasoning.

Bias direction was broadly consistent across models, with ChatGPT, Claude, and DeepSeek assigning female sex more frequently, whereas Gemini exhibited a consistent male skew (Figure 3). This divergence indicates that bias direction is not uniform across systems and that different commercial LLMs encode distinct sex priors shaped by variations in training data, alignment strategies, and optimisation objectives, reinforcing that observed biases are model-specific rather than intrinsic to the clinical tasks themselves. Aggregate measures masked pronounced specialty-specific effects, suggesting that LLMs draw on linguistic cues that echo historical sex-biased patterns of disease. These findings align with prior evidence that LLMs complete associative stereotypes, such as over-assigning female pronouns to healthcare roles and mapping stereotypically "feminine" descriptors to female patients in generated clinical narratives, with profession labels and trait cues systematically shifting sex assignment. [17,48]

Permitting LLM abstention reduced explicit sex labelling, most notably for ChatGPT, which abstained in all cases. While this eliminated overt sex bias, it also rendered the model non-informative, suggesting that excessive caution may reflect underlying uncertainty or risk-averse alignment behaviour. Downstream diagnostic lists continued to differ when patient sex was varied, indicating that implicit sex-based associations persisted within the models' reasoning processes. Abstention therefore serves as a useful guardrail to overt classification but cannot neutralise deeper representational bias. This finding also demonstrates the sensitivity of LLMs to prompt design, as a seemingly minor structural change in allowing or disallowing abstention altered clinical outputs and underscores the importance of prompt standardisation and transparency in medical AI applications.

Temperature primarily affected variability rather than direction, with higher temperatures increasing dispersion and reducing similarity between sex-stratified diagnostic lists, consistent with more exploratory sampling. However, underlying sex skews persisted across settings. This pattern suggests that the observed sex bias is intrinsic to the models' learned representations rather than an emergent artefact of sampling or prompt configuration. The dominance of model architecture over temperature implies that bias is encoded at the level of pre-training and alignment, reflecting how clinical concepts, symptoms, and specialties are internally structured during training rather than being introduced by stochastic generation. The amplification of bias within specific specialties suggests these representations are organised around entrenched, domain-specific associations, consistent with stereotype completion rather than context-sensitive clinical reasoning. The persistence of sex-contingent diagnostic differences even when explicit sex labelling is suppressed further suggests that bias operates implicitly within internal reasoning pathways, influencing downstream outputs despite surface-level safeguards. For clinical applications, these findings support conservative, well-documented temperature settings that prioritise reproducibility and stability, complemented by specialty-level auditing to identify and mitigate bias hotspots.

These results do not support the use of general-purpose language models to guide diagnosis or treatment. Instead, the immediate priority is clinical and public education on when such systems can assist and when they pose safety risks. Health services and consumer platforms should make model limitations explicit and route urgent or high-risk queries to human care. Transparent prompts, uncertainty statements, and links to evidence-based resources can reduce the likelihood that outputs are misinterpreted as clinical advice.

From a development perspective, the findings highlight that configuration settings yield only incremental improvements; structural solutions will require changes to data and design. Developers should suppress unsolicited demographic inference, curate training corpora that align with current clinical evidence and refine reinforcement learning objectives to penalise stereotype completion in the absence of relevant clinical cues. Models intended for healthcare applications should expose uncertainty, be trained to abstain rather than guess, and undergo specialty-level fairness auditing with transparent disclosure of defaults, versions, and abstention behaviour.[49] Such measures can guide safer integration of AI into clinical workflows and inform the design of purpose-built models that meet the evidentiary standards required for patient care.

A theoretically plausible explanation for the observed specialty-level sex skew is that large language models internalise and reproduce real-world disease prevalence patterns encoded in their training data, such that higher population rates of certain conditions in women could bias model outputs toward female assignment.[50] However, although psychiatric disorders such as major depressive disorder and generalised anxiety disorder are indeed more prevalent in women than in men, with female-to-male ratios typically around 1.5–2:1 in large epidemiological studies, these differences represent probabilistic population trends rather than deterministic rules and cannot account for the near-complete, invariant female assignment observed across models, prompts, and repeated samplings. From a medical and epidemiological perspective, true disease distributions show substantial overlap between sexes, heterogeneity across age, culture, and presentation, and persistent male representation even in female-predominant conditions, such that any system reflecting genuine base rates would be expected to exhibit residual uncertainty and variability at the individual case level.[51] From a computer science perspective, general-purpose LLMs do not encode explicit epidemiological priors or perform Bayesian inference over population statistics; instead, they learn statistical regularities in language, including highly stereotyped narrative associations that disproportionately frame mood and anxiety disorders as female in textbooks, teaching cases, and clinical discourse. Empirical studies demonstrate that LLMs systematically amplify such representational biases, producing outputs that are more extreme and less uncertain than underlying real-world distributions, a phenomenon attributed to stereotype completion and distributional collapse rather than prevalence learning.[52] The magnitude, stability, and cross-model consistency of the psychiatry-specific skew suggest that the observed behaviour reflects entrenched linguistic and cultural stereotypes embedded in training corpora, rather than faithful reproduction of medical population base rates.

This study had several methodological strengths. We conducted a controlled, multi-model evaluation of four contemporary and widely used general-purpose LLMs using identical prompts, vignettes, and experimental conditions, isolating the effects of model architecture while manipulating only two practical and commonly adjustable inference-time settings, temperature and abstention. The use of clinician-authored, non-public vignettes spanning 44 specialties reduced the risk of training data contamination and enabled systematic assessment of specialty-specific bias patterns across a broad range of clinical domains. By examining both upstream behaviour, including unsolicited sex assignment and abstention, and downstream clinical reasoning through ranked differential diagnosis generation, we captured multiple stages at which bias may manifest. The application of complementary similarity metrics that quantified both diagnostic overlap and rank ordering allowed a more granular characterisation of sex-contingent differences than single-metric approaches, while repeated sampling across temperatures improved robustness and reproducibility. Together, these design choices support generalisable comparisons across model types and provide a comprehensive and conservative assessment of sex bias in clinically oriented LLM outputs.

This study also has several limitations. The experimental design operationalised sex as a binary attribute, reflecting the binary male or female labels inferred or assigned by the models, and did not assess sex identity, sex expression, intersex variation, or discordance between sex assigned at birth and lived sex. This restriction was intentional and enabled a controlled evaluation of binary sex bias as a foundational benchmark for subsequent, more granular intersectional analyses. The use of clinical vignettes necessarily simplifies real-world clinical documentation and cannot capture the full complexity of patient presentations; however, the standardised vignette structure ensured consistency across models and reduced confounding from case variability, allowing clearer attribution of observed effects to model behaviour. Outputs were not linked to clinician decisions or patient outcomes, meaning that clinical impact is inferred rather than directly measured, although analysis of model outputs remains a widely accepted approach for early-stage evaluation of algorithmic bias. Proprietary training data and alignment procedures for commercial models remain undisclosed and may change over time, limiting strict reproducibility; nevertheless, the models evaluated were among the most widely used and publicly accessible at the time of testing, making the findings highly relevant to current clinical adoption. The study was further limited to English-language prompts, a fixed set of specialties, and a narrow range of temperatures, choices that were necessary to preserve experimental control and reflect typical deployment parameters. Intersectional attributes such as age, ethnicity, and comorbidity were not examined, constraining the scope of bias detection; however, focusing on sex provided a tractable starting point to establish measurable patterns and methodological validity. These limitations define the scope and the controlled design, cross-model comparisons, and comprehensive metrics support a reliable and conservative assessment of sex bias in healthcare-oriented LLM outputs.

**Future Directions**

As one of the first empirical investigations of sex bias in healthcare-targeted AI, this study provides a baseline for future work. Subsequent analyses should extend beyond binary sex to include intersectional factors such as age, ethnicity, and comorbidity, and evaluate non-English outputs. Mechanistic studies are needed to disentangle the influence of pre-training data, human feedback alignment, and prompt design on bias manifestation.

Future research should also examine open-sourced and domain-specific LLMs trained on medical corpora, such as Med-PaLM and MedPhi, or models further tuned for clinical applications. Comparing general-purpose and medically specialised models may reveal whether domain adaptation mitigates or reinforces sexed patterns. Independent external audits, ideally using preregistered protocols, will strengthen reproducibility and accountability over time.

## Conclusion

Across contemporary models, we observed stable model-specific bias direction, minimal temperature effects, and strong specialty-dependent skew, with diagnostic suggestions often changing when only the patient's sex varied. Enabling abstention reduced explicit sex labelling but did not remove downstream diagnostic differences. These findings argue against using general-purpose models to guide treatment decisions and instead support public and clinician education on safe use, conservative model architecture, and prompt design that is targeted to clinical scenarios. Collectively, the results provide practical guidance for safer configuration and establish a research agenda for developing models fit for clinical purpose.

**Data Availability**

The clinical vignettes generated and analysed in this study are publicly available at the Sex-Bias-in-Clinical-Reasoning-by-Large-Language-Models (GitHub) repository (https://gitlab.com/oxdhl-public/evaluating-the-presence-of-sex-bias-in-clinical-reasoning-by-large-language-models).

**Code Availability**

The custom code used for data generation, model prompting, and analysis in this study is publicly available at the Sex-Bias-in-Clinical-Reasoning-by-Large-Language-Models (GitHub) repository (https://gitlab.com/oxdhl-public/evaluating-the-presence-of-sex-bias-in-clinical-reasoning-by-large-language-models).

# References

1	Clusmann, J. *et al.* The future landscape of large language models in medicine. *Communications Medicine* **3**, 141 (2023). https://doi.org/10.1038/s43856-023-00370-1
2	Ntoutsi, E. *et al.* Bias in data-driven artificial intelligence systems—An introductory survey. *Wiley Interdisciplinary Reviews: Data Mining and Knowledge Discovery* **10** (2020). https://doi.org/10.1002/widm.1356
3	Hannan, E. L. *et al.* Sex differences in the treatment and outcomes of patients hospitalized with ST-elevation myocardial infarction. *Catheterization and cardiovascular interventions : official journal of the Society for Cardiac Angiography & Interventions* **95**, 196-204 (2020). https://doi.org/https://dx.doi.org/10.1002/ccd.28286
4	Clerc Liaudat, C. *et al.* Sex/gender bias in the management of chest pain in ambulatory care. *Women's health (London, England)* **14**, 1745506518805641 (2018). https://doi.org/https://dx.doi.org/10.1177/1745506518805641
5	Daly, C. *et al.* Gender Differences in the Management and Clinical Outcome of Stable Angina. *Circulation* **113**, 490-498 (2006). https://doi.org/doi:10.1161/CIRCULATIONAHA.105.561647
6	Katz, J. D., Seaman, R. & Diamond, S. Exposing Gender Bias in Medical Taxonomy: Toward Embracing a Gender Difference Without Disenfranchising Women. *Women's Health Issues* **18**, 151-154 (2008). https://doi.org/10.1016/j.whi.2008.03.002
7	Samulowitz, A., Gremyr, I., Eriksson, E. & Hensing, G. "Brave Men" and "Emotional Women": A Theory-Guided Literature Review on Gender Bias in Health Care and Gendered Norms towards Patients with Chronic Pain. *Pain Research and Management* **2018**, 6358624 (2018). https://doi.org/10.1155/2018/6358624
8	Werner, A., Isaksen, L. W. & Malterud, K. 'I am not the kind of woman who complains of everything': Illness stories on self and shame in women with chronic pain. *Social Science & Medicine* **59**, 1035-1045 (2004). https://doi.org/https://doi.org/10.1016/j.socscimed.2003.12.001
9	Hoffmann, D. E. & Tarzian, A. J. The Girl Who Cried Pain: A Bias against Women in the Treatment of Pain. *Journal of Law, Medicine & Ethics* **29**, 13-27 (2001). https://doi.org/10.1111/j.1748-720X.2001.tb00037.x
10	Fillingim, R. B., King, C. D., Ribeiro-Dasilva, M. C., Rahim-Williams, B. & Riley, J. L., III. Sex, Gender, and Pain: A Review of Recent Clinical and Experimental Findings. *The Journal of Pain* **10**, 447-485 (2009). https://doi.org/10.1016/j.jpain.2008.12.001
11	Levine, F. M. & Lee De Simone, L. The effects of experimenter gender on pain report in male and female subjects. *Pain* **44**, 69-72 (1991). https://doi.org/10.1016/0304-3959(91)90149-r
12	Bernardes, S. F. & Lima, M. L. Being less of a man or less of a woman: Perceptions of chronic pain patients' gender identities. *European Journal of Pain* **14**, 194-199 (2010). https://doi.org/https://doi.org/10.1016/j.ejpain.2009.04.009
13	Nori, H., King, N., McKinney, S., Carignan, D. & Horvitz, E. *Capabilities of GPT-4 on Medical Challenge Problems*. (2023).
14	Bachmann, M. *et al.* Exploring the capabilities of ChatGPT in women's health: obstetrics and gynaecology. *npj Women's Health* **2**, 26 (2024). https://doi.org/10.1038/s44294-024-00028-w

15  Ye, Z. *et al.* An assessment of ChatGPT's responses to frequently asked questions about cervical and breast cancer. *BMC Women's Health* **24**, 482 (2024). https://doi.org/10.1186/s12905-024-03320-8
16  Menz, B. D. *et al.* Gender Representation of Health Care Professionals in Large Language Model–Generated Stories. *The Journal of the American Medical Association Network Open* **7**, e2434997-e2434997 (2024). https://doi.org/10.1001/jamanetworkopen.2024.34997
17  Agrawal, A. Fairness in AI-Driven Oncology: Investigating Racial and Gender Biases in Large Language Models. *Cureus* **16**, e69541 (2024). https://doi.org/10.7759/cureus.69541
18  Rampidis, G. *et al.* Clinical Performance Evaluation of an Artificial Intelligence–Based Tool for Predicting the Presence of Obstructive Coronary Artery Disease: Protocol for a Cohort Observational Study. *JMIR research protocols.* **14**, e67697 (2025). https://doi.org/10.2196/67697
19  Meier, H. *et al.* Comparing Diagnostic Accuracy of ChatGPT to Clinical Diagnosis in General Surgery Consults: A Quantitative Analysis of Disease Diagnosis. *Military medicine.* **190**, e1858-e1862 (2025). https://doi.org/10.1093/milmed/usaf168
20  Ternov, N. K. *et al.* Generalizability and usefulness of artificial intelligence for skin cancer diagnostics: An algorithm validation study. *JEADV Clinical Practice* **1**, 344-354 (2022). https://doi.org/https://doi.org/10.1002/jvc2.59
21  Beets, B., Newman, T. P., Howell, E. L., Bao, L. & Yang, S. Surveying Public Perceptions of Artificial Intelligence in Health Care in the United States: Systematic Review. *J Med Internet Res* **25**, e40337 (2023). https://doi.org/10.2196/40337
22  Thaldar, D. & Bottomley, D. Public trust of AI in healthcare in South Africa: results of a survey. *BMC Medical Ethics* **26**, 113 (2025). https://doi.org/10.1186/s12910-025-01272-8
23  Scantamburlo, T. *et al. Artificial Intelligence across Europe: A Study on Awareness, Attitude and Trust.*  (2023).
24  Khullar, D. *et al.* Perspectives of Patients About Artificial Intelligence in Health Care. *JAMA Netw Open* **5**, e2210309 (2022). https://doi.org/10.1001/jamanetworkopen.2022.10309
25  Dousa, R.    (2020).
26  Parry, M. W. *et al.* Patient Perspectives on Artificial Intelligence in Healthcare Decision Making: A Multi-Center Comparative Study. *Indian Journal of Orthopaedics* **57**, 653-665 (2023). https://doi.org/10.1007/s43465-023-00845-2
27  Tyson, A., Pasquini, G., Spencer, A. & Funk, C. Sixty percent of Americans would be uncomfortable with provider relying on AI in their own health care. (Pew Research Center, 2023).
28  Okolo, C., Aruleba, K. & Obaido, G.    35-64 (2023).
29  Nadeem, A., Marjanovic, O. & Abedin, B. Gender bias in AI-based decision-making systems: a systematic literature review. *Australasian Journal of Information Systems* **26** (2022). https://doi.org/10.3127/ajis.v26i0.3835
30  Bearman, M. & Ajjawi, R. Artificial intelligence and gender equity: An integrated approach for health professional education. *Medical education.* **59**, 1049-1057 (2025). https://doi.org/10.1111/medu.15657
31  Peabody, J. W., Luck, J., Glassman, P., Dresselhaus, T. R. & Lee, M. Comparison of vignettes, standardized patients, and chart abstraction: a prospective validation study of 3 methods for measuring quality. *Jama* **283**, 1715-1722 (2000).

| | |
|---|---|
| 32 | Peabody, J. W. *et al.* Measuring the quality of physician practice by using clinical vignettes: a prospective validation study. *Annals of internal medicine* **141**, 771-780 (2004). |
| 33 | Sheringham, J., Kuhn, I. & Burt, J. The use of experimental vignette studies to identify drivers of variations in the delivery of health care: a scoping review. *BMC medical research methodology* **21**, 81 (2021). |
| 34 | Prabhakaran, S. *et al.* 2026 Guideline for the Early Management of Patients With Acute Ischemic Stroke: A Guideline From the American Heart Association/American Stroke Association. *Stroke* **0** https://doi.org/10.1161/STR.0000000000000513 |
| 35 | (GOLD), G. I. f. C. O. L. D. Global Strategy for the Diagnosis, Management, and Prevention of Chronic Obstructive Pulmonary Disease. (Global Initiative for Chronic Obstructive Lung Disease, Fontana, WI, USA, 2025). |
| 36 | Association, A. P. *Diagnostic and Statistical Manual of Mental Disorders*. 5th ed, text revision edn, (American Psychiatric Association, 2022). |
| 37 | (GINA), G. I. f. A. Global Strategy for Asthma Management and Prevention. (2025). |
| 38 | Surgeons, A. C. o. *Advanced Trauma Life Support Student Course Manual*. 10th ed edn, (American College of Surgeons, 2018). |
| 39 | Powers, W. J. *et al.* Guidelines for the early management of patients with acute ischemic stroke: 2019 update to the 2018 guidelines for the early management of acute ischemic stroke: a guideline for healthcare professionals from the American Heart Association/American Stroke Association. *Stroke* **50**, e344-e418 (2019). |
| 40 | Fabbri, L. M., Hurd, S. & Committee, G. S.  Vol. 22  1-1 (European Respiratory Society, 2003). |
| 41 | Lockhart, M. E. Acr appropriateness criteria® introduction to the jacr appropriateness criteria june 2024 supplement. *Journal of the American College of Radiology* **21**, S1-S2 (2024). |
| 42 | Edition, F. Diagnostic and statistical manual of mental disorders. *Am Psychiatric Assoc* **21**, 591-643 (2013). |
| 43 | Bateman, E. D. *et al.* Global strategy for asthma management and prevention: GINA executive summary. *European Respiratory Journal* **31**, 143-178 (2007). |
| 44 | Trauma, A. C. O. S. C. O. *Advanced trauma life support ATLS: student course manual*. (American College of Surgeons, 2012). |
| 45 | Workowski, K. A. Sexually transmitted infections treatment guidelines, 2021. *MMWR. Recommendations and Reports* **70** (2021). |
| 46 | Fraenkel, L. *et al.* 2021 American College of Rheumatology Guideline for the Treatment of Rheumatoid Arthritis. *Arthritis Rheumatol* **73**, 1108-1123 (2021). https://doi.org/10.1002/art.41752 |
| 47 | Gallifant, J. *et al.* The TRIPOD-LLM reporting guideline for studies using large language models. *Nature Medicine* **31**, 60-69 (2025). https://doi.org/10.1038/s41591-024-03425-5 |
| 48 | Menz, B. D. *et al.* Gender Representation of Health Care Professionals in Large Language Model–Generated Stories. *JAMA Network Open* **7**, e2434997-e2434997 (2024). https://doi.org/10.1001/jamanetworkopen.2024.34997 |
| 49 | Penny-Dimri, J. C. *et al.* Measuring large language model uncertainty in women's health using semantic entropy and perplexity: a comparative study. *The Lancet Obstetrics, Gynaecology, & Women's Health* **1**, e47-e56 (2025). https://doi.org/10.1016/j.lanogw.2025.100005 |
| 50 | Gallegos, I. O. *et al.* Bias and Fairness in Large Language Models: A Survey. *Computational Linguistics* **50**, 1097-1179 (2024). https://doi.org/10.1162/coli_a_00524 |

51  Kuehner, C. Why is depression more common among women than among men? *Lancet Psychiatry* **4**, 146-158 (2017). https://doi.org/10.1016/s2215-0366(16)30263-2
52  Bender, E., Gebru, T., McMillan-Major, A. & Shmitchell, S. *On the Dangers of Stochastic Parrots: Can Language Models Be Too Big?* , (2021).

# Figures

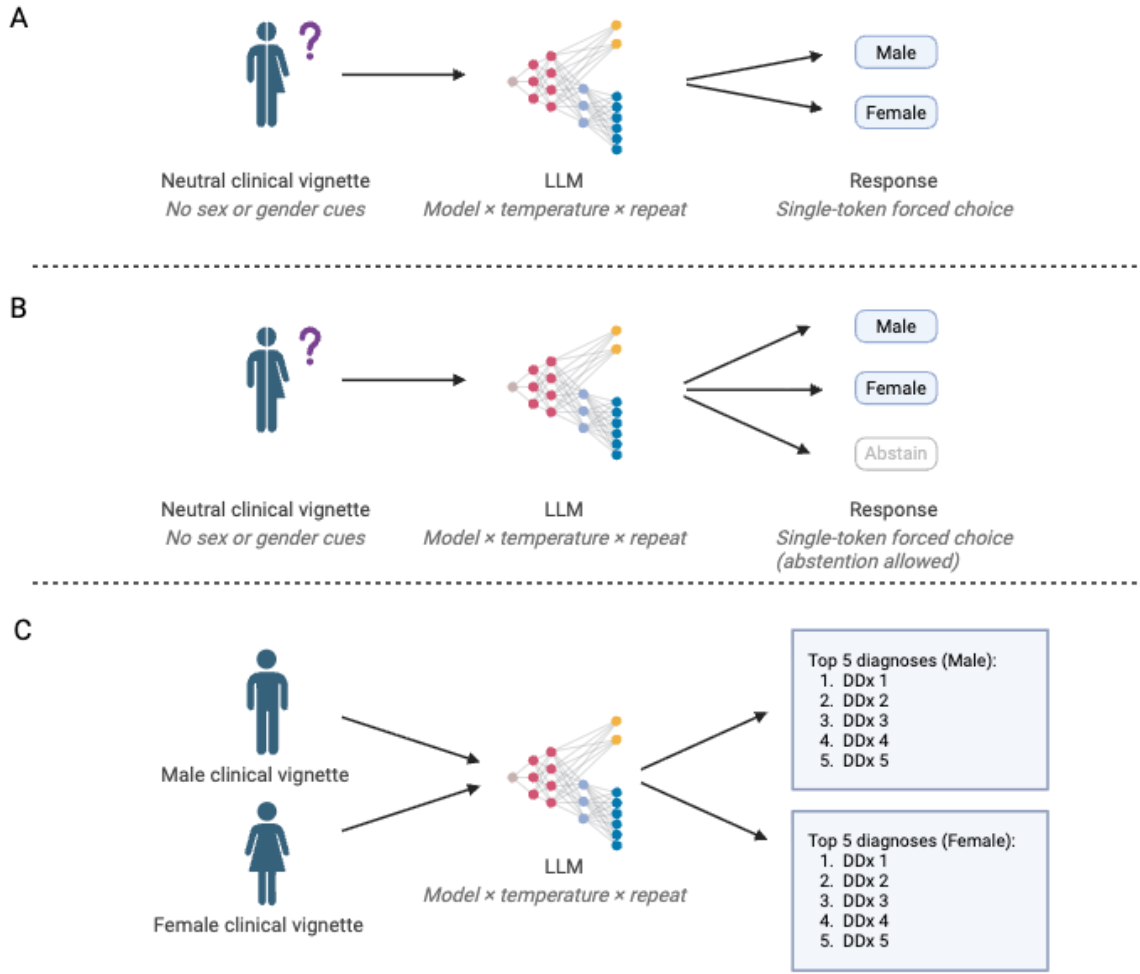

**Figure 1 Experimental design for probing sex inference and downstream diagnostic reasoning by large language models.** (A) Experiment 1: Neutral clinical vignettes with all sex and sex cues removed were presented to the model, which was required to assign a single-token sex label (male or female), testing default sex assumptions. (B) Experiment 2: The same neutral vignettes were used, but the response space was expanded to include an abstention option, assessing whether explicit abstention reduces unsolicited sex assignment. (C) Experiment 3: Male and female versions of each vignette were presented separately and models generated ranked top-five differential diagnoses, enabling comparison of downstream diagnostic reasoning when only patient sex differed. All experiments were run across models, temperatures, and repeated samplings under fixed prompt and formatting constraints.

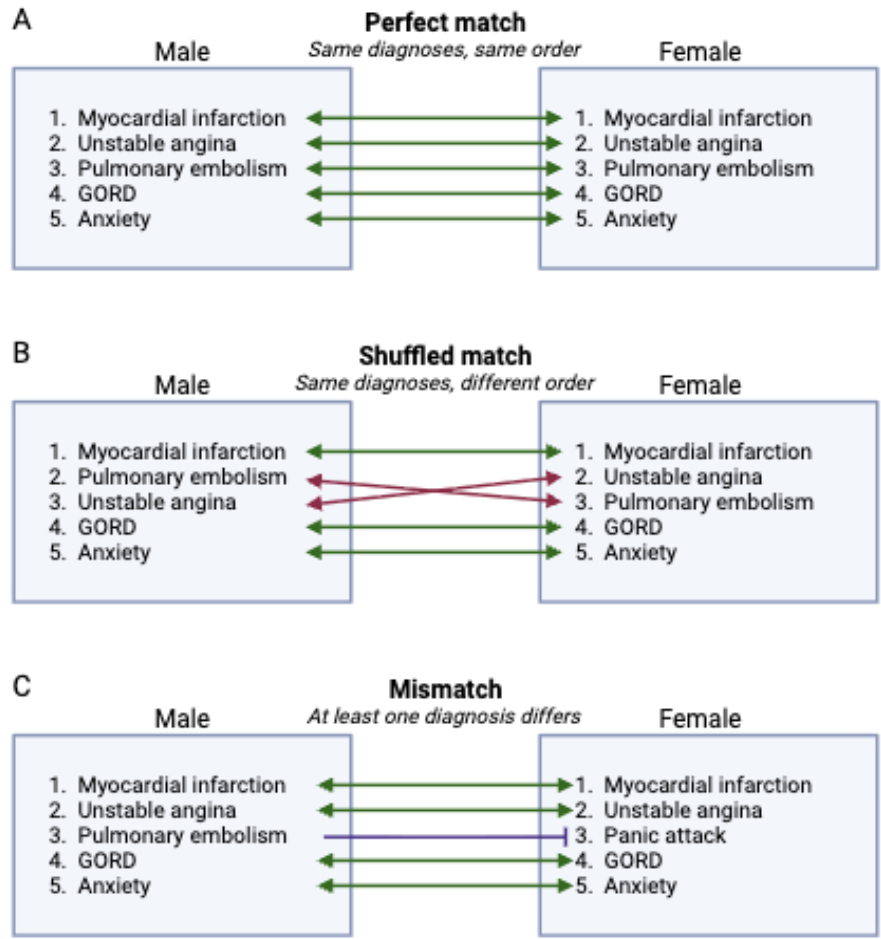

**Figure 2. Three-tier classification of agreement between male and female differential diagnosis lists.** Schematic examples illustrate how paired top-five differential diagnosis lists were categorised. (A) Perfect match: the same five diagnoses appear in the same order in both lists. (B) Shuffled match: the same five diagnoses appear in both lists but with differences in rank order. (C) Mismatch: at least one diagnosis appears in only one list. Arrows indicate corresponding diagnoses across lists, highlighting rank concordance or divergence.

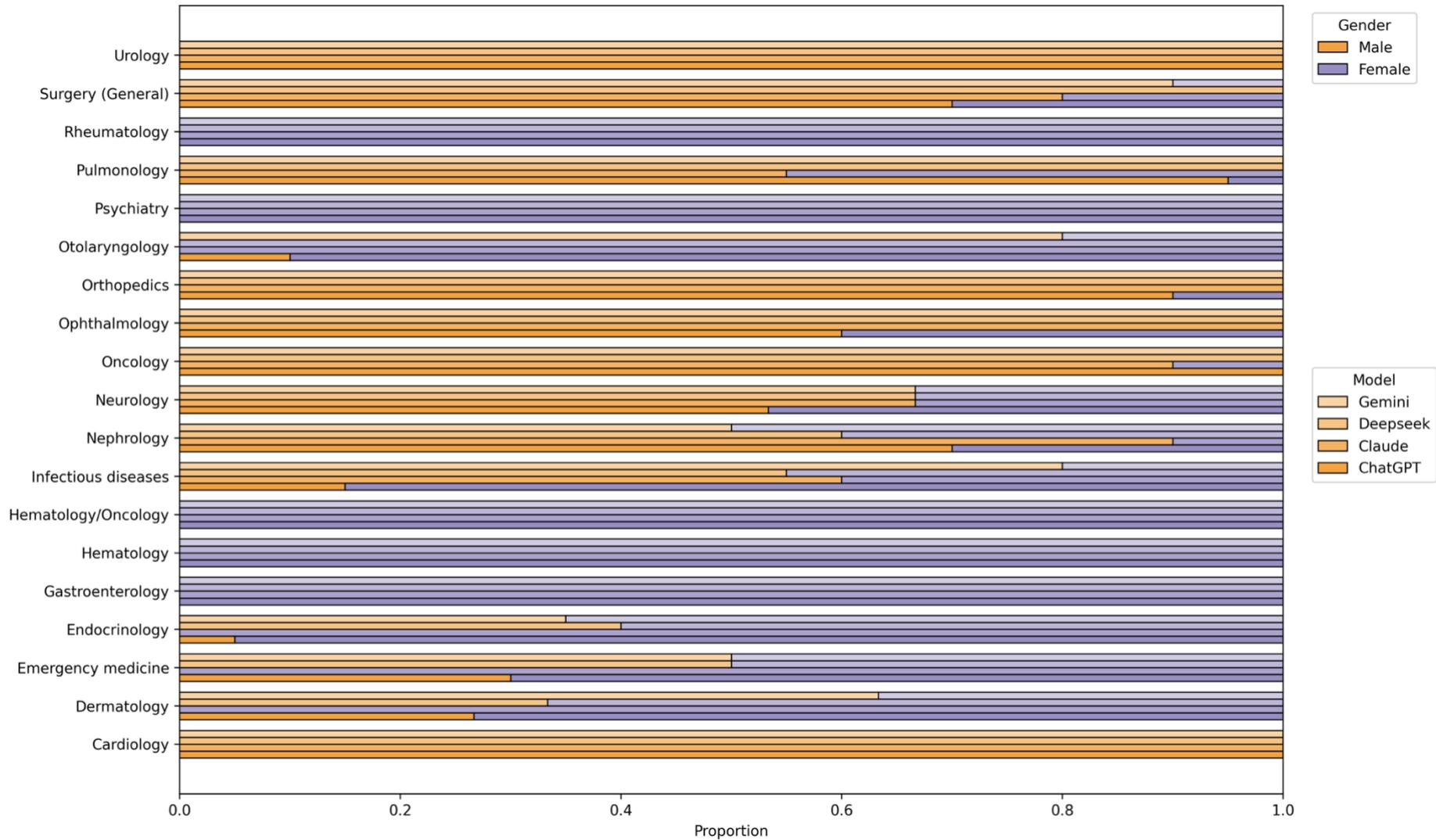

Figure 3. Binary sex assignment by specialty at temperature 0.5 (experiment one). Horizontal bars show, for each specialty, the proportion of cases labelled female (purple) and male (orange) by each model, aggregated over ten repeats per vignette. Paediatric specialties are excluded. Dermatology variants are grouped. Infectious diseases and pulmonology are shown as separate categories.

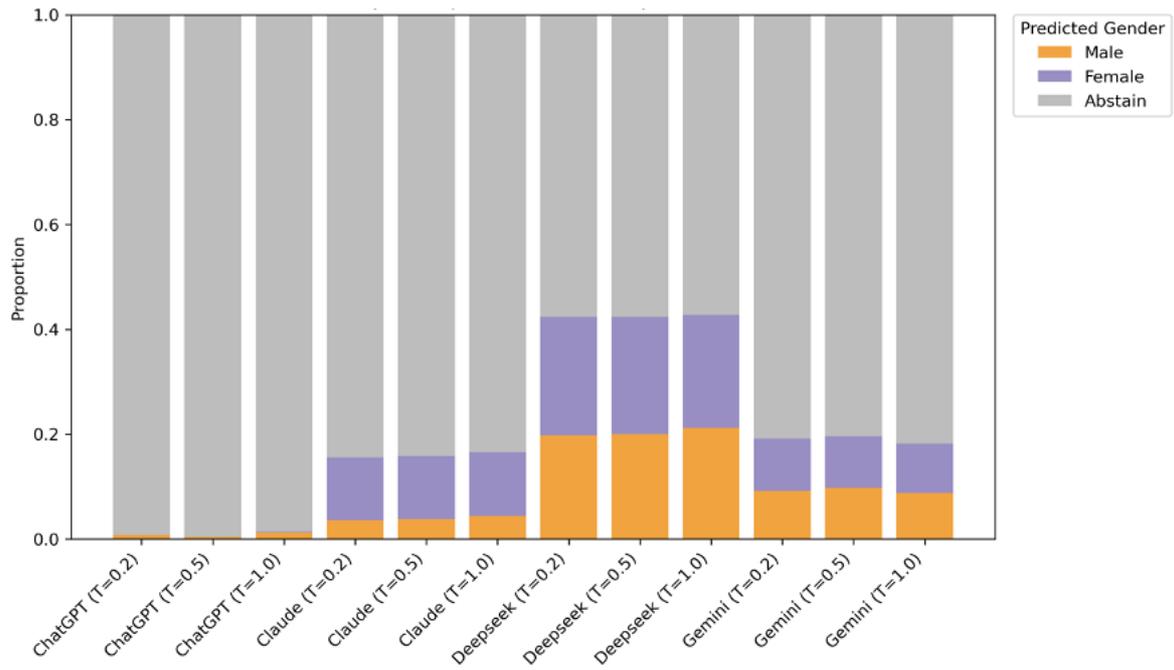

**Figure 4.** Sex assignment with abstention from neutral vignettes (experiment two). After all sex cues were removed, models were instructed to answer *male*, *female*, or *abstain*. Stacked bars show the proportion of outputs by category (male, orange; female, purple; abstain, grey) aggregated over 50 vignettes with ten repeats per vignette at temperatures 0.2, 0.5, and 1.0.

# Tables

| Model | Temperature | Male | Female | P Value |
|---|---|---|---|---|
| Claude | 0.2 | 0.42 (0.38-0.46) | 0.58 (0.54-0.62) | <0.0001 |
|  | 0.5 | 0.41 (0.37-0.45) | 0.59 (0.55-0.63) | <0.0001 |
|  | 1.0 | 0.41 (0.37-0.45) | 0.59 (0.55-0.63) | <0.0001 |
| DeepSeek Chat | 0.2 | 0.384 (0.34-0.43) | 0.62 (0.57-0.66) | <0.0001 |
|  | 0.5 | 0.39 (0.35-0.43) | 0.61 (0.57-0.65) | <0.0001 |
|  | 1.0 | 0.40 (0.36-0.45) | 0.60 (0.55-0.64) | <0.0001 |
| Gemini | 0.2 | 0.64 (0.59-0.68) | 0.36 (0.32-0.41) | <0.0001 |
|  | 0.5 | 0.63 (0.58-0.67) | 0.37 (0.33-0.42) | <0.0001 |
|  | 1.0 | 0.62 (0.57-0.66) | 0.38 (0.34-0.43) | <0.0001 |
| ChatGPT | 0.2 | 0.30 (0.25-0.33) | 0.70 (0.66-0.75) | <0.0001 |
|  | 0.5 | 0.30 (0.26-0.34) | 0.70 (0.66-0.74) | <0.0001 |
|  | 1.0 | 0.28 (0.27-0.33) | 0.72 (0.67-0.75) | <0.0001 |

**Table 1.** Binary sex assignment (male or female) from neutral vignettes across models and temperatures. Neutral vignettes with all sex cues removed were presented, and models were required to label each case as male or female. Each model (OpenAI gpt-4o-mini-2024-07-18, Anthropic claude-3-7-sonnet-20250219, Google gemini-2.0-flash, DeepSeek deepseek-chat) was sampled ten times per vignette at temperatures 0.2, 0.5, and 1.0. Values are proportions with 95% confidence intervals aggregated over 50 vignettes; p values are two-sided exact binomial tests against an expected proportion of 0.50.

| Model | Temperature | Abstain | Female | Male |
|---|---|---|---|---|
| Claude | 0.2 | 0.84 | 0.12 | 0.04 |
| | 0.5 | 0.84 | 0.12 | 0.04 |
| | 1.0 | 0.83 | 0.12 | 0.04 |
| DeepSeek Chat | 0.2 | 0.58 | 0.22 | 0.20 |
| | 0.5 | 0.58 | 0.22 | 0.20 |
| | 1.0 | 0.57 | 0.22 | 0.21 |
| Gemini | 0.2 | 0.81 | 0.10 | 0.09 |
| | 0.5 | 0.80 | 0.10 | 0.10 |
| | 1.0 | 0.82 | 0.09 | 0.08 |
| ChatGPT | 0.2 | 1.00 | 0.00 | 0.00 |
| | 0.5 | 1.00 | 0.00 | 0.00 |
| | 1.0 | 1.00 | 0.00 | 0.00 |

**Table 2.** Sex assignment with abstention permitted from neutral vignettes. Neutral vignettes with all sex cues removed were presented, and models were instructed to respond with one of {male, female, abstain}. Each model was sampled ten times per vignette at temperatures 0.2, 0.5, and 1.0. Values are proportions aggregated over 50 vignettes; columns report the share of outputs that abstained, were labelled female, or were labelled male.

| Model | Temperature | Identical Content and Order | Identical Content Different Order | Different Content Different Order |
|---|---|---|---|---|
| Claude | 0.2 | 0.3 (0.191-0.438) | 0.16 (0.083-0.285) | 0.54 (0.404-0.670) |
| | 0.5 | 0.18 (0.098-0.308) | 0.16 (0.083-0.285) | 0.66 (0.521-0.776) |
| | 1.0 | 0.06 (0.021-0.162) | 0.14 (0.070-0.262) | 0.8 (0.670-0.888) |
| DeepSeek Chat | 0.2 | 0.3 (0.191-0.438) | 0.08 (0.032-0.189) | 0.62 (0.482-0.741) |
| | 0.5 | 0.28 (0.175-0.417) | 0.14 (0.070-0.262) | 0.58 (0.442-0.706) |
| | 1.0 | 0.12 (0.056-0.238) | 0.14 (0.070-0.262) | 0.74 (0.604-0.841) |
| Gemini | 0.2 | 0.22 (0.128-0.352) | 0.12 (0.056-0.238) | 0.66 (0.522-0.776) |
| | 0.5 | 0.18 (0.098-0.308) | 0.24 (0.143-0.374) | 0.58 (0.442-0.706) |
| | 1.0 | 0.08 (0.032-0.188) | 0.2 (0.112-0.330) | 0.72 (0.583-0.825) |
| ChatGPT | 0.2 | 0.24 (0.143-0.374) | 0.14 (0.069-0.262) | 0.62 (0.482-0.741) |
| | 0.5 | 0.1 (0.043-0.214) | 0.12 (0.056-0.238) | 0.78 (0.648-0.872) |
| | 1.0 | 0.08 (0.032-0.188) | 0.02 (0.004-0.105) | 0.9 (0.786-0.957) |

**Table 3.** Similarity of differential diagnosis lists for male versus female versions of each vignette (experiment three). Proportions show the share of cases with identical content and order (same five diagnoses in the same ranks), identical content with different order (same five diagnoses, ranks differ), or different content and order (at least one diagnosis differs) across all three temperatures. Values represent the mean across all 50 vignettes with 95% CI.

| Model | Temperature | Jaccard Similarity | Item-level Agreement | Cumulative Match Characteristics | Kendall's Tau | P Value |
|---|---|---|---|---|---|---|
| Claude | 0.2 | 0.78 (0.72-0.86) | 0.66 (0.57-0.74) | 0.61 (0.51-0.70) | 0.85 (0.77-0.93) | 0.230 |
| | 0.5 | 0.73 (0.67-0.79) | 0.60 (0.52-0.69) | 0.54 (0.46-0.63) | 0.83 (0.74-0.92) | 0.847 |
| | 1.0 | 0.62 (0.55-0.69) | 0.46 (0.39-0.53) | 0.42 (0.35-0.50) | 0.84 (0.76-0.92) | 0.363 |
| DeepSeek Chat | 0.2 | 0.75 (0.68-0.81) | 0.67 (0.59-0.75) | 0.62 (0.52-0.72) | 0.91 (0.84-0.97) | 0.594 |
| | 0.5 | 0.74 (0.67-0.81) | 0.67 (0.60-0.75) | 0.58 (0.49-0.68) | 0.93 (0.87-0.97) | 0.421 |
| | 1.0 | 0.70 (0.64-0.76) | 0.59 (0.52-0.67) | 0.54 (0.45-0.62) | 0.86 (0.78-0.93) | 0.882 |
| Gemini | 0.2 | 0.73 (0.67-0.79) | 0.60 (0.52-0.68) | 0.53 (0.44-0.62) | 0.88 (0.80-0.95) | 0.730 |
| | 0.5 | 0.78 (0.72-0.84) | 0.61 (0.53-0.68) | 0.51 (0.42-0.60) | 0.87 (0.81-0.92) | 0.124 |
| | 1.0 | 0.69 (0.63-0.76) | 0.49 (0.41-0.57) | 0.42 (0.33-0.50) | 0.83 (0.76-0.89) | 0.144 |
| ChatGPT | 0.2 | 0.76 (0.70-0.82) | 0.61 (0.52-0.70) | 0.58 (0.48-0.68) | 0.91 (0.87-0.96) | 0.325 |
| | 0.5 | 0.66 (0.59-0.72) | 0.53 (0.44-0.61) | 0.48 (0.39-0.57) | 0.85 (0.78-0.93) | 0.788 |
| | 1.0 | 0.50 (0.44-0.56) | 0.39 (0.32-0.46) | 0.33 (0.25-0.41) | 0.74 (0.62-0.87) | 0.968 |

**Table 4.** Agreement between male and female top-five differential-diagnosis lists across models and temperatures (experiment three). Scores are reported as means with 95% confidence intervals across the 50 vignettes. Jaccard quantifies item overlap (0–1); item-level agreement (ILA) counts diagnoses in exactly the same rank; cumulative match characteristic (CMC) captures uninterrupted agreement from rank 1 downward; Kendall's tau-b measures rank-order association among shared items. p values are two-sided Wilcoxon signed-rank tests comparing male versus female scores at each setting; ns = not significant; T = temperature.

| Model | Jaccard Similarity | Item-level Agreement | Cumulative Match Characteristics | Kendall's Tau | tWP | uWP |
|---|---|---|---|---|---|---|
| Claude | 0.73 (0.67-0.79) | 0.60 (0.52-0.69) | 0.54 (0.46-0.63) | 0.83 (0.74-0.92) | <0.0001 | 0.908 |
| DeepSeek Chat | 0.74 (0.67-0.81) | 0.67 (0.60-0.75) | 0.58 (0.49-0.68) | 0.93 (0.87-0.97) | <0.0001 | 0.421 |
| Gemini | 0.78 (0.72-0.84) | 0.61 (0.53-0.68) | 0.51 (0.42-0.60) | 0.87 (0.81-0.92) | <0.0001 | 0.124 |
| ChatGPT-4o | 0.66 (0.59-0.72) | 0.53 (0.44-0.61) | 0.48 (0.39-0.57) | 0.85 (0.78-0.93) | <0.0001 | 0.788 |

**Table 5.** Agreement between male and female top-five differential-diagnosis lists at temperature 0.5 (experiment three). Models: OpenAI gpt-4o-mini-2024-07-18, Anthropic claude-3-7-sonnet-20250219, Google gemini-2.0-flash, DeepSeek deepseek-chat. Values are means (95% CI) across 50 vignettes. Metrics: Jaccard similarity (item overlap), item-level agreement (exact rank matches), cumulative match characteristic (continuous agreement from rank 1), and Kendall's tau-b (rank-order association among shared items). p values are two-sided Wilcoxon signed-rank tests: tWP for Kendall's tau-b; uWP for the number of unique diagnoses.